\documentclass[journal]{IEEEtran}
\usepackage[utf8]{inputenc}
\usepackage{color}
\usepackage{xcolor}
\usepackage{array}
\usepackage{verbatim}
\usepackage{float}
\usepackage{amsmath}
\usepackage{amsthm}
\usepackage{amssymb}
\usepackage{graphicx}
\usepackage{longtable}
\usepackage{multirow}
\usepackage{booktabs}
\usepackage[unicode=true,
bookmarks=false,
breaklinks=false,pdfborder={0 0 1},colorlinks=false]
{hyperref}
\hypersetup{
	colorlinks,bookmarksopen,bookmarksnumbered,citecolor=blue,urlcolor=blue}
\usepackage{cite}

\usepackage{lipsum}
\usepackage{mathtools}
\usepackage{cuted}
\providecommand{\tabularnewline}{\\}
\usepackage{algorithmic}
\usepackage{longtable}

\floatstyle{ruled}
\newfloat{algorithm}{tbp}{loa}
\providecommand{\algorithmname}{Algorithm}
\floatname{algorithm}{\protect\algorithmname}

\makeatletter
\let\oldforeign@language\foreign@language
\DeclareRobustCommand{\foreign@language}[1]{%
	\lowercase{\oldforeign@language{#1}}}

\let\oldforeign@language\foreign@language
\DeclareRobustCommand{\foreign@language}[1]{%
	\lowercase{\oldforeign@language{#1}}}

\ifCLASSINFOpdf
\else
\fi

\@ifundefined{showcaptionsetup}{}{%
	\PassOptionsToPackage{caption=false}{subfig}}
\usepackage{subfig}

\usepackage{balance}

\ifCLASSINFOpdf
\else
\fi

\ifCLASSINFOpdf
\else
\fi

\def\ps@IEEEtitlepagestyle{%
	\def\@oddhead{\parbox[t][\height][t]{\textwidth}{\centering \scriptsize
			Personal use of this material is permitted. Permission from the author(s) and/or copyright holder(s), must be obtained for all other uses. Please contact us and provide details if you believe this document breaches copyrights.\\
			\noindent\makebox[\linewidth]{}
		}\hfil\hbox{}}%
	\def\@evenhead{\scriptsize\thepage \hfil \leftmark\mbox{}}%
	\def\@oddfoot{\parbox[t][\height][l]{\textwidth}{
			\vspace{-20pt}{\rule{\textwidth}{0.4pt}}\\ \footnotesize{\bf{\footnotesize\textcolor{red}{A. M. Ali, A. Gupta, and H. A. Hashim, "Deep Reinforcement Learning for Sim-to-Real Policy Transfer of VTOL-UAVs Offshore Docking Operations," Applied Soft Computing, vol. 162, pp. 111843, 2024.}}}doi: \href{https://doi.org/10.1016/j.asoc.2024.111843}{10.1016/j.asoc.2024.111843}\\\\
			\noindent\makebox[\linewidth]
		}\hfil\hbox{}}%
	\def\@evenfoot{\MYfooter}}

\makeatother
\begin{document}
	\bstctlcite{IEEEexample:BSTcontrol}

\title{Deep Reinforcement Learning for Sim-to-Real Policy Transfer of VTOL-UAVs Offshore Docking Operations}

\author{Ali M. Ali, Aryaman Gupta, and Hashim A. Hashim% <-this % stops a space
	\thanks{This work was supported in part by the National Sciences and Engineering Research Council of Canada (NSERC), under the grants RGPIN-2022-04937.}
	\thanks{A. M. Ali, A. Gupta, and H. A. Hashim are with the Department of Mechanical
		and Aerospace Engineering, Carleton University, Ottawa, ON, K1S-5B6,
		Canada (e-mail: hhashim@carleton.ca).}
}

% \markboth{IEEE TRANSACTIONS ON INTELLIGENT TRANSPORTATION SYSTEMS, \today}{Hashim \MakeLowercase{\textit{et al.}}: Landmark and IMU Data Fusion: Systematic Convergence Geometric Nonlinear Observer for SLAM and Velocity Bias}

% \markboth{}{Hashim \MakeLowercase{\textit{et al.}}: Nonlinear Filter for Simultaneous Localization and Mapping on a Matrix Lie Group using IMU and Feature Measurements}

\maketitle
\begin{abstract}
This paper proposes a novel Reinforcement Learning
(RL) approach for sim-to-real policy transfer of Vertical Take-Off
and Landing Unmanned Aerial Vehicle (VTOL-UAV). The proposed approach
is designed for VTOL-UAV landing on offshore docking stations in maritime
operations. VTOL-UAVs in maritime operations encounter limitations
in their operational range, primarily stemming from constraints imposed
by their battery capacity. The concept of autonomous landing on a
charging platform presents an intriguing prospect for mitigating these
limitations by facilitating battery charging and data transfer. However,
current Deep Reinforcement Learning (DRL) methods exhibit drawbacks,
including lengthy training times, and modest success rates. In this
paper, we tackle these concerns comprehensively by decomposing the
landing procedure into a sequence of more manageable but analogous
tasks in terms of an approach phase and a landing phase. The proposed
architecture utilizes a model-based control scheme for the approach
phase, where the VTOL-UAV is approaching the offshore docking station.
In the Landing phase, DRL agents were trained offline to learn the
optimal policy to dock on the offshore station. The Joint North Sea
Wave Project (JONSWAP) spectrum model has been employed to create
a wave model for each episode, enhancing policy generalization for
sim2real transfer. A set of DRL algorithms have been tested through
numerical simulations including value-based agents and policy-based
agents such as Deep \textit{Q}
Networks (DQN) and Proximal Policy Optimization (PPO) respectively.
The numerical experiments show that the PPO agent can learn complicated
and efficient policies to land in uncertain environments, which in
turn enhances the likelihood of successful sim-to-real transfer.
\end{abstract}


\section{Introduction}\label{sec1}

\subsection{Motivation}
\IEEEPARstart{V}{ertical} Take-off and Landing Unmanned Aerial Vehicles
(VTOL-UAVs) have been a standard part of offshore operations over
the last few decades. VTOL-UAVs provide faster, safer, and more cost-efficient
solutions for the inspection and maintenance of offshore operations
for instance as in oil and gas facilities. The rapid growth of Micro-Electro-Mechanical
Systems (MEMSs) contributed to the development of the VTOL-UAVs, however,
there are still several challenges to be solved \cite{hashim2019special,zielinski20213d}.
Flight endurance is a persisting challenge which limits the VTOL-UAV
usage to only temporary missions in offshore operations \cite{IEEEsurvey2016UAVs}.
Lithium polymer batteries are often used for UAV energy supply. Although
Lithium batteries capacity has been improved over the last decade,
they still provide a limited energy supply \cite{battery2013}.

Offshore ocean operations can be accomplished through charging pads
or docking stations with the capability to recharge or exchange lithium
polymer batteries \cite{applications2019,docking2019ASME}. Various
designs of docking stations have been discussed in the literature,
ranging from wireless charging pads to designs based on mechanisms
of completely replacing the UAV battery \cite{IEEETransElectroincs2023}.
Wireless charging stations free from human-based wired plugging/unplugging
or replacing the batteries have been considered utilizing complicated
mechanisms \cite{IEEETranswirelessdesgin2022}. In \cite{Wirelessdesginone,WirelessdesginTwo,Wirelessdesginthree,Wirelessdesginfour,Wirelessdesginfive}
various wireless charging designs have been proposed with the merit
of the lightweight and compact design of the charging pad. However,
wireless charging stations are sensitive to misalignment leading to
a reduction of power transmission efficiency. As an illustration,
in the case of \cite{Wirelessdesginsix}, the 200 Watt model exhibits
a radial misalignment tolerance distance of merely 20 millimeters
(mm). In \cite{Wirelessdesginseven} a wireless charging coupling
structure was devised using copper tubes at both the transmitter and
receiver. The transmission efficiency in \cite{wirelessdesgineight}
was 78\%--80\% within a radial misalignment of 32 cm to 44 cm. It
becomes apparent that deploying a meticulous docking strategy can
markedly decrease power loss during the charging phase.

The well-defined dynamic model of VTOL-UAVs motivated the active research
community to formulate model-based controllers to decide either on
static or moving platform docking operations \cite{hashim2023uwb,ali2024mpc}.
In \cite{HamelIEEETransdocking2012}, the VTOL-UAV was presumed to
come with a basic sensor package comprising at least a camera and
an inertial measurement unit, where the authors provided a rigorous
analysis of system stability of the proposed controller based on Lyapunov
functions. Vision-based nonlinear model predictive control (MPC) followed
by boundary layer sliding mode was proposed in \cite{NonlinearmpcICRA2020},
where simulations and experiments showed successful landing in a moving
platform with wind disturbances. Multi-sensor fusion techniques for
localization have been utilized in the docking operations as in \cite{sensorfusionone2017,sensorfusiontwo2018,sensorfusionthree2019}
for more precise docking. Despite the success of the model-based controllers,
the architecture is handcrafted to solve specific situations and usually
requires significant engineering effort and large computational power
as in the case of MPC. 

In recent years, the robotics community has extensively embraced Reinforcement
Learning (RL) algorithms for controlling complex single-robot and
multi-robot systems \cite{yan2023immune,arulkumaran2017deep,nguyen2020deep,abouheaf2023online,kumar2023deep},
establishing end-to-end policies, and in turn encompassing perception
and control strategies \cite{garcia2015comprehensive}. RL's foundation
lies in emulating human trial-and-error learning, with agents acquiring
knowledge through rewards obtained from their actions under various
conditions. This reliance on rewards entails a significant number
of episodes, thus exposing RL's learning constraints concerning time
and experience variability in real-world applications. The training
of RL agents on real robots has been found to yield limited successes
\cite{levine2015learning,levine2018learning,levine2016end}. In contrast,
contemporary RL algorithms have demonstrated remarkable performance
when trained with ample amounts of inexpensive synthetic data from
virtual experiments. Consequently, a prevalent strategy in learning
schemes involves conducting agent training within simulated environments
that closely approximate real-world conditions \cite{peng2018sim,tobin2017domain,tan2018sim,julian2020scaling,golemo2018sim,christiano2016transfer,rusu2017sim}.

In \cite{hwangbo2017control,koch2019reinforcement} Deep Reinforcement
Learning (DRL) has been employed in conventional control problems
related to VTOL-UAV, encompassing attitude control, set-point tracking,
hovering, and disturbance recovery. Utilizing a low-resolution camera
facing downward, a Deep-Q Network (DQN) identified a marker and successfully
executed a landing in \cite{bicer2019vision}. A Deep Deterministic
Policy Gradient (DDPG) algorithm was implemented to guide a VTOL-UAV
for landing on a moving platform \cite{bicer2019vision,polvara2018toward,polvara2019autonomous}.
In \cite{kooi2021inclined}, the study addresses the challenging task
of landing a VTOL-UAV on inclined surfaces, where conventional control
methods are not applicable due to the non-equilibrium nature of the
final state. Unfortunately, current DRL docking schemes exhibit drawbacks,
including lengthy training times, and modest success rates. The lengthy
training times arise from learning complicated three-dimensional (3D)
navigation policies, while the primary cause of modest success rates
lies in the disparity between the simulated environment and real-world
conditions.

\paragraph*{Contributions} Inspired by the limitations
found in the literature on DRL-based docking of VTOL-UAV schemes,
the paper contributions can be outlined as follows:
\begin{itemize}
	\item Unlike \cite{hwangbo2017control,koch2019reinforcement},
	the docking task of VTOL-UAV is decomposed into a sequence of more
	manageable but analogous phases. The approach phase and the landing
	phase reduce the training time. Also, the proposed learning architecture
	learns the optimal policy to land on a docking platform considering
	uncertainty and unknown movement in the docking landing phase; in
	contrast to \cite{hwangbo2017control,koch2019reinforcement} which
	focus on stable and static docking platform.
	\item A novel generalized policy for sim-to-real transfer
	based on domain randomization has been provided to increase the success
	rates in landing operations. The DRL agents learn using a virtual
	environment subjected to a randomized disturbance describing the hydrodynamic
	waves represented by the Joint North Sea Wave Project (JONSWAP) spectrum
	model.
	\item A set of policy-based and value-based DRL agents
	have been tested with respect to learning stability against the randomized
	disturbance. 
\end{itemize}
\paragraph*{Organization}The remaining part of the paper is organized
as follows: Section \ref{sec:Preliminaries} presents preliminaries
on DRL. The Problem formulation is stated in Section \ref{sec:Problem-Formulation}
along with the proposed DRL agents and the randomized disturbance
virtual environment. Section \ref{sec:Numerical-Experiments} illustrates
the performance of the proposed learning agents scheme through numerical
simulations. Finally, Section \ref{sec:Conclusions-=000026-Future}
concludes the work.

\section{Preliminaries\label{sec:Preliminaries}}

\subsection{Deep Reinforcement Learning }

Let $\mathbb{R}$ be the set of real numbers and $\mathbb{Z}^{+}$
be the set of non-negative integers. A finite Markov decision process
is a stochastic control process in discrete time, represented as a
4-tuple such that $\text{MDP}=(S,A,p,r)$ where:
\begin{itemize}
	\item $S$ is set of a finite number of states.
	\item $A$ is set of a finite number of actions.
	\item $r:S\times A\rightarrow\mathbb{R}:r(s^{\prime}|s,a)$ is the expected
	instantaneous reward obtained upon transitioning from state $s\in S$
	to a new state $s^{\prime}\in S$, due to action $a\in A$ where $|$
	denotes the notation of the conditional probability. 
\end{itemize}
The reward can be written as:
\begin{flalign*}
	r(s,a) & =\sum_{r\in\mathbb{R}}r\sum_{s^{\prime}\in S}p(s^{\prime},r|s,a)
\end{flalign*}

\begin{itemize}
	\item $p:S\times A\times S\rightarrow[0,1]$ is the state transition probability
	where action $a$ in state $s\in S$ at time $t\in\mathbb{Z}^{+}$
	leads to state $s^{\prime}\in S$ at time $t+1.$ $p$ can be defined
	as:
\end{itemize}
\begin{flalign*}
	p(s^{\prime},r|s,a) & =\sum_{r\in R}p(s^{\prime},r|s,a)
\end{flalign*}
In reinforcement learning, the fundamental aim is formalized through
a reward signal transferred from the environment to the agent at each
time step $r_{t}\in\mathbb{R}.$ The agent's goal is to maximize the
total cumulative reward received over the long run, which means not
maximizing the immediate reward. The agent endeavors to devise a policy
for decision-making that maximizes the cumulative discounted reward
anticipated in the future. The expected discounted return $R$ at
time $t\in\mathbb{Z}^{+}$ is defined as follows:
\begin{align}
	R_{t} & =\mathbb{E}[\sum_{k=0}^{\infty}\gamma^{k}r_{t+k+1}]\label{eq:3}
\end{align}
where $0\leq\gamma\leq1$ is the discount factor and $k$ is the time
step. Let $V_{\pi}(s)$ be the value of a state $s$ under a policy
$\pi$; and it represents the expected return when starting in the
state $s$ such that:
\begin{equation}
	V_{\pi}(s)=\mathbb{E}_{\pi}[\sum_{k=0}^{\infty}\gamma^{k}r_{t+k+1}|s_{t}=s]\label{eq:4}
\end{equation}
where $\mathbb{E}_{\pi}[\cdot]$ represents the expected value of
a random variable given that the agent adheres policy $\pi$. In this
manner, the value of taking action $a$ in state $s$ under a policy
$\pi$, denoted by $Q_{\pi}(s,a)$ and it is termed the action value
function. It defines the excepted return starting from $s$ taking
action $a$ and following a policy $\pi$ such that:
\begin{equation}
	Q_{\pi}(s,a)=\mathbb{E}_{\pi}[\sum_{k=0}^{\infty}\gamma^{k}r_{t+k+1}|s_{t}=s,a_{t}=a]\label{eq:5}
\end{equation}
from Eq. \eqref{eq:4} and \eqref{eq:5} the advantage function $A_{\pi}(s,a)$
can be defined as follows:
\begin{equation}
	A_{\pi}(s,a)=Q_{\pi}(s,a)-V_{\pi}(s)\label{eq:6}
\end{equation}
Given an estimate of $Q_{\pi}(s_{t},a_{t})$, it is possible to enhance
the policy by, for instance, acting greedily based on $Q_{\pi}(s_{t},a_{t})$
for instance:
\begin{equation}
	\pi^{\prime}(s_{t})=\text{argmax}_{a}Q_{\pi}(s_{t},a).\label{eq:7}
\end{equation}
where $\pi'(s_{t})$ is the improved policy. The conventional $Q$
Learning algorithm conducts policy evaluation through temporal difference
estimation and adjusts the $Q$ estimation according to the Bellman
Eq. \cite{10.5555/3312046} as follows:
\begin{multline}
	Q_{\pi}^{\text{new}}(s_{t},a_{t})=Q_{\pi}^{\text{old}}(s_{t},a_{t})+\alpha(r_{t}+\\
	\gamma\max_{a}Q_{\pi}^{old}(s_{t+1},a)-Q_{\pi}^{old}(s_{t},a_{t}))\label{eq:10}
\end{multline}
where $\alpha\in\mathbb{R}$ refers to a step size. The conventional
standard tabular Q-learning method is challenged by the curse of dimensionality,
rendering computationally intractable in scenarios where the number
of states and actions grows substantially, a common occurrence across
various practical applications. An alternative approach leveraging
artificial neural networks for estimating the $Q$ function, giving
rise to the DQN algorithm as initially proposed in \cite{mnih2013playing}.
Algorithms that aim to estimate the $Q$ function and subsequently
make decisions based on the current estimation fall into the category
of value-based algorithms, while if we directly learn the optimal
policy $\pi_{\theta}^{*}$ through a neural network that maximizes
a given objective function that will be called policy based algorithms. 

\section{Problem Formulation \label{sec:Problem-Formulation}}

\begin{figure}[h]
	\begin{centering}
		\includegraphics[scale=0.5]{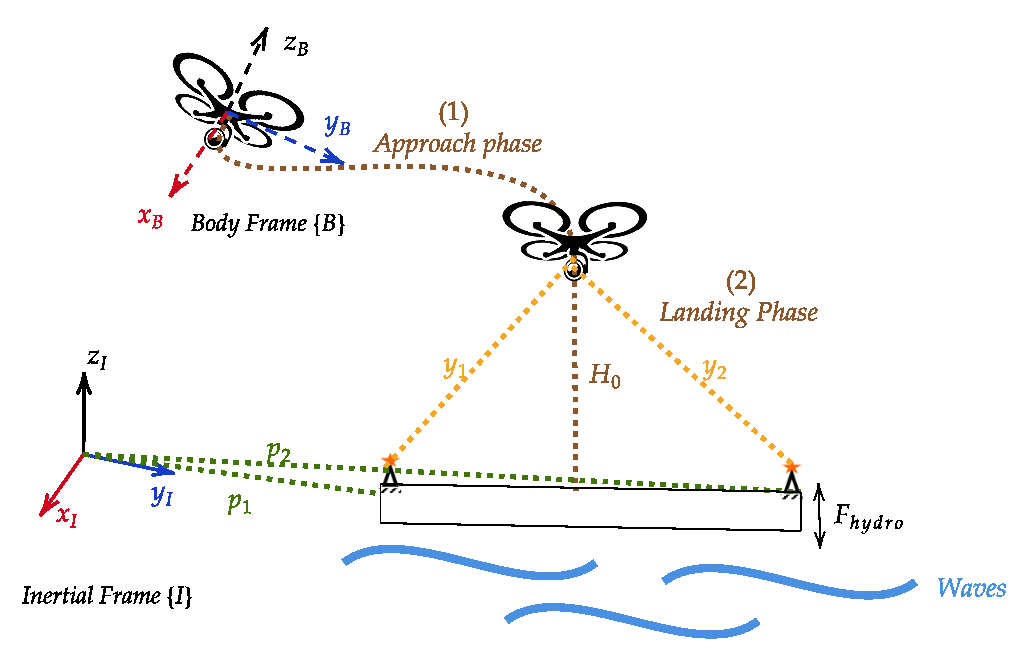}
		\par\end{centering}
	\caption{VTOL-UAV landing task on an offshore landing pad, where $\{\mathcal{B}\}$
		refers to the body-fixed frame and $\{\mathcal{I}\}$ is the Inertial-frame.\textcolor{black}{{}
			The distances $\{y_{1},y_{2}\}$ are from the VTOL-UAV to the landmarks
			attached to the docking station to be used in the approach phase.
			The distances $\{p_{1},p_{2}\}$ are from the inertial frame to the
			corners of the landing pad. For more information about the approach
			phase using landmark please refer to \cite{hashim2023exponentially},
			\cite{hashim2023observer}.}}
	\label{fig:formulation}
\end{figure}

This Section aims to formulate DRL agents to learn the optimal policy
that enables a safe and accurate landing on an offshore docking station
subjected to disturbance. Define $\{\mathcal{I}\}$ as the inertial
frame coordinates and $\{\mathcal{B}\}$ as the body frame coordinates.
Recall the mapping between a rotational matrix $R\in SO(3)$ to Euler
angles $\zeta=[\phi,\theta,\psi]^{\top}\in\mathbb{R}^{3}$ ($R\rightarrow\zeta$)
and vice versa $\zeta\rightarrow R$ where $SO(3)$ refers to the
Special Orthogonal Group of order 3 (for more information visit \cite{hashim2019special}):
\begin{equation}
	R_{\zeta}=\left[\begin{array}{ccc}
		c_{\psi}c_{\theta} & c_{\psi}s_{\phi}s_{\theta}-c_{\phi}s_{\psi} & s_{\phi}s_{\psi}+c_{\phi}c_{\psi}s_{\theta}\\
		c_{\theta}s_{\psi} & c_{\phi}c_{\psi}+s_{\phi}s_{\psi}s_{\theta} & c_{\phi}s_{\psi}s_{\theta}-c_{\psi}s_{\phi}\\
		-s_{\theta} & c_{\theta}s_{\phi} & c_{\phi}c_{\theta}
	\end{array}\right].\label{eq:1-2}
\end{equation}
The dynamics of a rigid body subject to external forces applied to
its center of mass and articulated in Newton-Euler formalism is delineated
as follows:

\begin{equation}
	\sum F=m\dot{V},\hspace{1cm}F,V\in\{\mathcal{I}\}\label{eq:18}
\end{equation}

\begin{equation}
	\sum M=J\dot{\Omega}+\Omega\times J\Omega\hspace{1cm}M,\Omega\in\{\mathcal{\mathcal{B}}\}\label{eq:19}
\end{equation}
where $V=[\begin{array}{ccc}
	v_{x} & v_{y} & v_{z}\end{array}]^{\top}\in\mathbb{R}^{3}$ is the UAV linear velocity vector in the inertial frame $\{\mathcal{I}\}$,
$\Omega=[\begin{array}{ccc}
	\Omega_{x} & \Omega_{y} & \Omega_{z}\end{array}]^{\top}\in\mathbb{R}^{3}$ describes UAV rotational velocity in the body frame $\{\mathcal{B}\}$,
$m$ is the total mass of the UAV, $J$ refers to the inertia matrix,
and $\times$ denotes the cross product between two vectors. $\sum F$
and $\sum M$ are the sum of the external forces and moments acting
on the VTOL-UAV. To facilitate the formulation of a learning framework,
our aim is to decompose the docking into two phases; the approach
phase and the landing phase as illustrated in Figure \ref{fig:formulation}.
The approach phase is a 3D navigation and tracking control problem,
which can be successfully addressed by the rich literature on such
problems (e.g., \cite{hashim2023observer,hashim2023exponentially}).
In \cite{hashim2023observer} and \cite{hashim2023exponentially},
the authors used known landmarks obtained through a vision-aided unit
(monocular or stereo camera) and an Inertial Measurement Unit (IMU).
This enabled the development of an observer-based controller with
exponential stability, facilitating effective 3D navigation and tracking
control for VTOL-UAVs. Utilizing model-based control schemes in the
approach phase is reasonable since the model of VTOL-UAV is known,
and there are no significant advantages in learning a complicated
policy using model-free DRL agents. Nevertheless, the adaptation of
the model-based control in the approach phase will provide a more
manageable learning task in the landing phase. The landing phase will
start when the approach phase is finished, where the VTOL-UAV will
navigate in 3D space and align with the docking station with a fixed
height denoted by $H_{0}$. The DRL agents learn the optimal policy
to drive the VTOL-UAV to land accurately with minimal impact force
against the disturbed docking station. Learning offline in a simulated
environment with a randomized disturbance will provide an optimal
generalized policy in terms of neural network mapping the current
state to the optimal action providing fast online solutions. 

\subsection{Environment}

This subsection aims to develop a mathematical framework for a virtual
environment representing the landing phase, tasked with training the
DRL agents offline. In the view of Eq. \eqref{eq:18}, the sum of
the external forces $\sum F$ can be written as follows:
\begin{equation}
	\sum F=\underbrace{\left[\begin{array}{c}
			0\\
			0\\
			-mg
		\end{array}\right]}_{\text{gravity}}+R_{\zeta}\underbrace{\left[\begin{array}{c}
			0\\
			0\\
			U_{1}
		\end{array}\right]}_{\text{total-thrust}}\label{eq:311}
\end{equation}
As such, from Eq. \eqref{eq:311} the vertical dynamics of the VTOL
UAV can be described by:
\begin{alignat}{1}
	z & \rightarrow\dot{x}_{1}=x_{2}\label{eq:31}\\
	\dot{z} & \rightarrow\dot{x}_{2}=\frac{-k_{fdz}}{m}x_{1}-g+U\label{eq:32}
\end{alignat}
where $x=[\begin{array}{cc}
	x_{1} & x_{2}\end{array}]^{\top},$ $x_{1}=z$ , $x_{2}=\dot{z}$ , $k_{fdz}$ denotes the aerodynamic
drag coefficient in the $z$ direction, and $U$ refers to the virtual
control direction acting on the direction of $z_{b}$ such that the
actual thrust $u=\frac{m}{c_{\phi}c_{\theta}}U$. Starting from initial
height $H_{0}$ and assuming that VTOL-UAV is perfectly aligned from
the approach phase lead to $c_{\phi}=c_{\theta}=1$, one finds the
stabilizing hover controller as follows:
\[
U_{0}=\frac{k_{fdz}}{m}x_{1}+g.
\]
The formulation of the DQN agents requires a discrete action space,
unlike the Proximal Policy Optimization (PPO) agent which can accommodate
continuous action space. The discrete action space is defined by $A_{d}=[U_{+},U_{0},U_{-}]$
where $U_{+}$ and $U_{-}$ refers to the set of control values with
a bigger and smaller values with respect to $U_{0}$, respectively.
The continuous action is defined by $A_{c}=\{U:U_{\min}\leq U\leq U_{\max}\}$
where $U_{\min}$ and $U_{\max}$ are the minimum and maximum values
of the virtual control input, respectively. The state space $S=\left[\begin{array}{cc}
	z-z_{w} & \dot{z}\end{array}\right]^{\top}\in\mathbb{R}^{2}$ is defined for both DQN and PPO agents, where $z_{w}$ is the vertical
displacement of the docking platform due to the $F_{\text{hydro}}$
to be defined subsequently. The dynamics in Eq. \eqref{eq:31} and
\eqref{eq:32} will be numerically integrated using the SciPy \cite{oliphant2007python}
Adams/backward differentiation formula methods in Python at each time
step. The docking station position at the origin is subjected to a
stochastic noise from the hydrodynamic waves $F_{hydro}$. The JONSWAP
spectrum is a widely utilized model in oceanography and engineering
for describing the energy distribution in ocean waves \cite{kumar2008spectral}.
It is characterized by a peak enhancement at a certain frequency and
a rapid decrease in energy at frequencies away from the peak. The
JONSWAP spectrum is described by:
\begin{equation}
	S(f)=\frac{\alpha_{w}g^{2}}{k_{w}^{4}}(f)^{-5}e^{(\frac{-5}{4}(\frac{f_{p}}{f})^{4})}\gamma_{w}^{e^{(\frac{-(f-f_{p})^{2}}{2\sigma^{2}f_{p}^{2}})}}\label{eq:33-1}
\end{equation}
where $f$ is the wave frequency, $S(f)$ represents the spectral
energy density at a wave frequency $f$, $\alpha_{w}$ is the Phillips
constant, which is typically set as $0.0081$ for deep water, $k_{w}$
is the von Karman constant set as 0.016 for deep water, $f_{p}$ is
the peak frequency at which the spectrum has the maximum energy, $\gamma_{w}$
is a dimensionless shape parameter that helps to define the shape
of the spectrum, and $\sigma$ is the spectral width parameter, which
influences the width of the spectrum. The Fourier transform $X(\omega)$
of the spectrum is obtained, followed by the generation of the time
domain wave using the inverse Fourier transform such that:
\begin{equation}
	X(f)=\intop_{-\infty}^{\infty}S(f)e^{-2\pi ft}df\label{eq:34}
\end{equation}
where the magnitude is given by $\vert X(\omega)\vert=\sqrt{Re(X(\omega))^{2}+X(f(\omega))}$
and the phase is defined by $\phi_{\omega}=\tan^{-1}(\frac{Im(X(\omega))}{Re(X(\omega))})$.
A random value $\phi_{\omega}=[0,2\pi]$ is used to generate a random
wave single for each episode. This results in a random time domain
signal that represents the desired wave such that $z_{w}(t)=\mathcal{F}^{-1}[X(\omega)]$,
where $\mathcal{F}^{-1}$ is the inverse Fourier transform and $z_{w}$
is the landing position disturbed by the hydrodynamic waves.

\subsection{Reward }

The formal definition of a reinforcement learning problem necessitates
the specification of the reward function. In this work, the reward
function is formulated to give a higher reward to the VTOL-UAV while
approaching the landing pad and minimizing the impact force. Define
$e_{p}=z-z_{w}$ and $e_{v}=\dot{z}-\dot{z}_{d}$ to be the error
in height and height rate, respectively. In order to maintain a soft
landing with minimal impact force, one can assign the required velocity
$\dot{z}_{d}$ to do so (e.g., exponential decay of the required velocity).
The proposed reward function can be defined as follows:
\begin{equation}
	R(s,a)=-k_{1}e_{p}-k_{2}e_{v}.\label{eq:33}
\end{equation}
where $k_{1}$ and $k_{2}$ are gain constants to give more weight
of $e_{p}$ or $e_{v}$ within the reward value.

\subsection{Value based Agents }

The fundamental concept underlying value-based reinforcement learning
algorithms is to iteratively update and estimate the action-value
function $Q$ in Eq. \eqref{eq:5} using the Bellman equation as follows:
\begin{equation}
	Q_{k+1}(s,a)=\mathbb{E}[R+\gamma\max_{a'}(s',a')|s,a]\label{eq:303}
\end{equation}
Conventional tabular value-based algorithms converge to the optimal
action value function, $Q_{k}$ → $Q^{*}$ as $i\rightarrow\infty$
\cite{sutton2018reinforcement}. In practice, tabular value-based
algorithms is computationally untraceable, due to the issue of dimensionality
arising from the vast number of possible values of states ($z$,$\dot{z}$).
Alternatively, it is typical to employ a function approximator to
estimate the action-value function, denoted as, $Q(s,a;\theta)\thickapprox Q^{*}(s,a)$. 

\subsubsection{DQN}

A neural network function approximator with parameters represented
by $\theta$ is referred to as a $Q$ network. In \cite{mnih2013playing},
a $Q$ network can be trained by minimizing a sequence of loss functions
$\mathcal{L}_{k}(\theta_{k})$ defined in Eq. \eqref{eq:34-1} that
changes at each iteration $k$.
\begin{figure*}[h]
	\begin{centering}
		\includegraphics[scale=0.6]{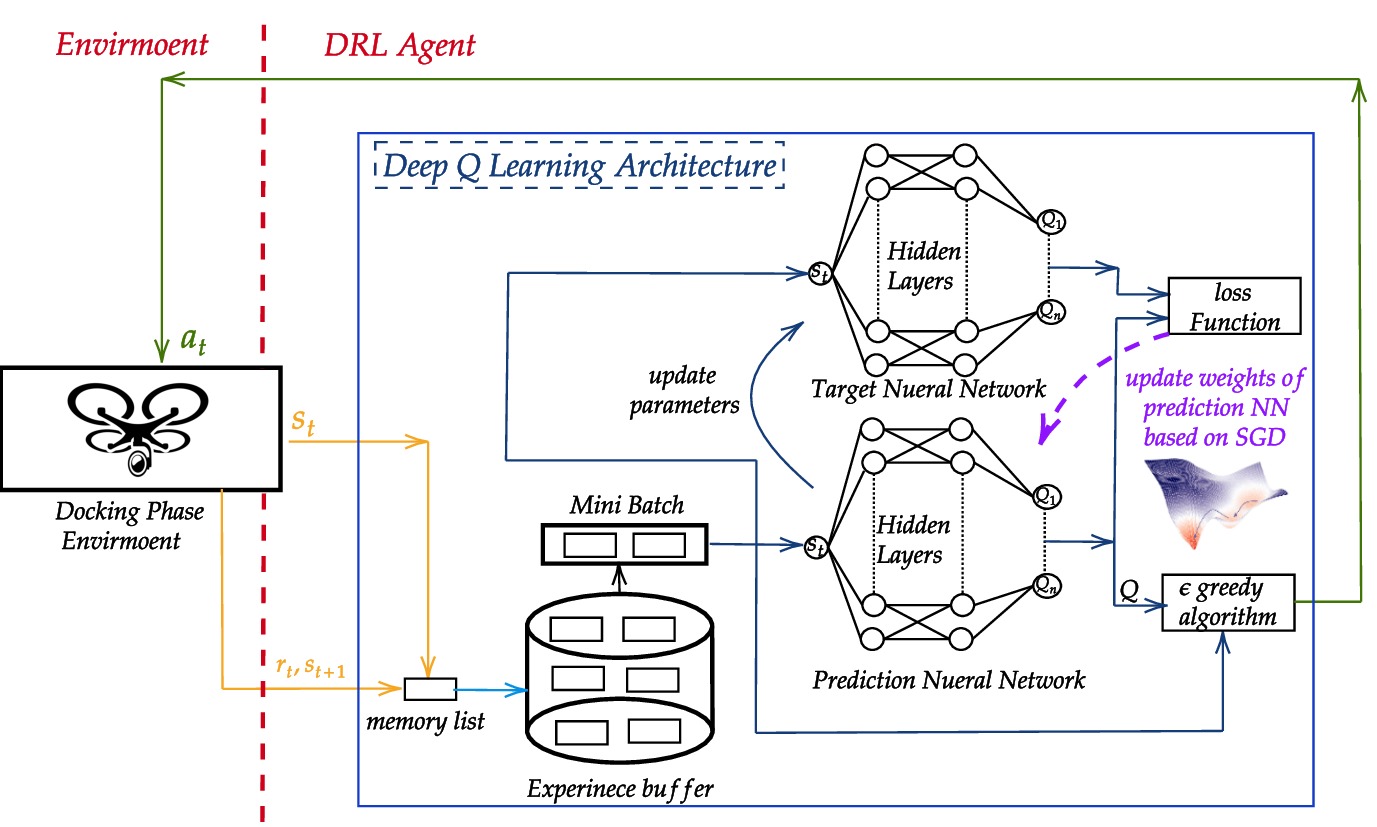}
		\par\end{centering}
	\caption{DQN Agent architecture interacting with the docking phase environment.}
	
	\label{fig:DQN}
\end{figure*}
\begin{algorithm}[H]
	\caption{DQN Docking Learning Algorithm}
	\label{Algo:1}
	
	\textbf{Input: }$M$, $C$, $F$, $\gamma$, $\epsilon_{0}$, $\epsilon_{f}$,
	$\alpha$, $f_{p}$, and $\gamma_{w}$.
	
	\textbf{Output: $\theta$}
	
	\textbf{Initialization:}
	\begin{enumerate}
		\item[{\footnotesize{}1:}]  Initialize $\theta_{\text{prediction}}$, and $\theta_{\text{target}}$.
	\end{enumerate}
	\textbf{for }every episode\textbf{ $=1,\cdots,M$ do}
	\begin{enumerate}
		\item[{\footnotesize{}5:}]  Initialize $s_{1}$ and $\theta_{\text{prediction}}=\text{\ensuremath{\theta_{\text{prediction}}(s_{1})}.}$
		\item[{\footnotesize{}6:}]  Compute $\vert X(\omega)\vert=\sqrt{Re(X(\omega))^{2}+X(f(\omega))}$
		from Eq. \eqref{eq:33-1} and \eqref{eq:34}.
		\item[{\footnotesize{}7:}]  Generate a random phase $\phi_{\omega}=[0,2\pi]$.
		\item[{\footnotesize{}8:}]  Generate the wave time signal from $z_{w}(t)=\mathcal{F}^{-1}[X(\omega)].$
	\end{enumerate}
	\textbf{\hspace{1cm}for }every time step\textbf{ $k=1,\cdots,k_{f}.$}
	\begin{enumerate}
		\item[{\footnotesize{}9:}]  Every $F$ Steps set $\theta_{\text{pediction}}=\theta_{\text{target}}$.
		\item[{\footnotesize{}10:}]  compute a random number $\zeta\sim\text{Uniform}(0,1)$.
	\end{enumerate}
	\textbf{\hspace{1.25cm}if} $\zeta>\epsilon$ then
	\begin{enumerate}
		\item[{\footnotesize{}11:}] \textbf{\hspace{0.5cm}} select $a_{k}=\text{argmax}_{a^{'}\in A(s_{t})}Q(s_{k},a',\theta_{\text{prediction}})$.
	\end{enumerate}
	\textbf{\hspace{1.25cm}else}
	\begin{enumerate}
		\item[{\footnotesize{}12:}] \textbf{\hspace{0.5cm}} select a random action $a_{k}\in A$.
	\end{enumerate}
	\textbf{\hspace{1.25cm}end if}
	\begin{enumerate}
		\item[{\footnotesize{}13:}]  Apply the action $a_{k}$.
		\item[{\footnotesize{}14:}]  Numerically integrate the dynamics in Eq. \eqref{eq:31} and \eqref{eq:32}
		using SciPy.
		\item[{\footnotesize{}15:}]  Store $(s_{k},a_{k},r_{k},s_{k+1})$ in $D$.
		\item[{\footnotesize{}16:}]  Sample a random mini batch of $(s_{j},a_{j},r_{j},s_{k+1})$ from
		$D.$
		\item[{\footnotesize{}17:}]  Set $Q_{j}=r_{j}+\gamma\max_{a'}Q(\phi_{j+1},a';\theta_{\text{target}})$
		\item[{\footnotesize{}18:}]  execute a gradient descent step to minimize\\
		$\mathcal{L}=(Q_{j}-Q(s_{j},a_{j};\theta_{\text{prediction}}))^{2}$
		\item[{\footnotesize{}19:}]  update $\theta\leftarrow\theta+\alpha\nabla_{\theta}J(\theta).$
	\end{enumerate}
	\textbf{\hspace{1cm}end for}
	
	\textbf{end for}
	\begin{enumerate}
		\item[{\footnotesize{}20:}]  Save $\theta_{\text{prediction}}$
	\end{enumerate}
\end{algorithm}
\begin{figure}[h]
	\begin{centering}
		\includegraphics[scale=0.42]{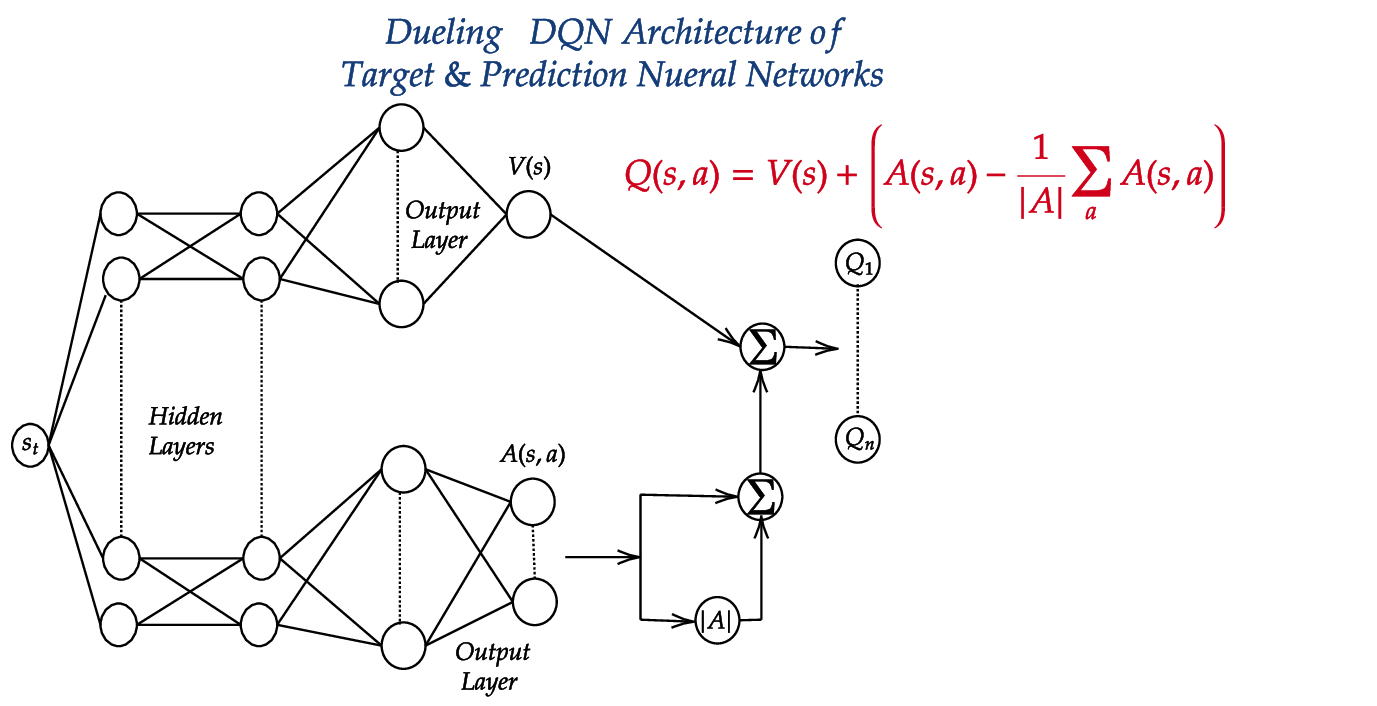}
		\par\end{centering}
	\caption{Dueling Architecture of Target \& Prediction Neural Networks.}
	\label{fig:dueling}
\end{figure}
\begin{equation}
	\mathcal{L}_{k}(\theta_{k})=(y_{k}-Q(s_{k},a_{k};\theta_{\text{prediction}}))^{2}\label{eq:34-1}
\end{equation}
where $y_{k}=r+\gamma\max_{a'}Q(s',a';\theta_{k-1})$ is the target
for time step $k$. The weights from the previous time step $\theta_{k}$
are held fixed when optimizing the loss function $\mathcal{L}_{k}(\theta_{k})$.
It is worth noting that the targets are contingent on the network
weights. This differs from the fixed targets employed in supervised
learning, which remain unchanged before the learning process commences.
Stochastic Gradient Descent (SGD) can be utilized to minimize the
loss function in \eqref{eq:34-1}. Using the experience replay technique,
the agent’s experiences at each time step will be a sequence of $(s_{k},a_{k},r_{k},s_{k+1})$
stored in memory $D$ with capacity $C$. The JONSWAP spectrum wave
model is embedded in the calculation of the dynamics of the environment.
Within the inner loop of the algorithm, $Q$ values updates and mini-batch
updates are applied to randomly sampled experiences from the stored
pool. Following the completion of experience replay, the agent chooses
and executes an action based on an $\epsilon$-greedy policy. The
details of the implementation of the algorithm are shown in Algorithm
\ref{Algo:1} and Figure \ref{fig:DQN}.

\paragraph*{Algorithm \ref{Algo:1} Discussion}

The algorithm begins by initializing essential components such as
the replay memory with capacity $C$, number of episodes $M$, update
frequency $F$, discount factor $\gamma$, initial exploration $\epsilon_{0}$,
final value $\epsilon_{f}$, learning rate $\alpha$ and other parameters
identifying the wave spectral energy in Eq. \eqref{eq:33-1} as $\alpha_{w}$,
$f_{p}$, and $\gamma_{w}$. It then proceeds to iterate through each
episode, computing the magnitude of the Fourier Transform in Eq. \eqref{eq:34}
and generating a random phase before deriving the time-domain wave
signal from the inverse Fourier transform. Within each time step,
the algorithm updates the prediction neural network every $F$ steps
to match the target network, selects an action based on the current
state and neural network predictions, and applies this action to the
system. It then stores the resulting transition in the replay memory
and samples a random mini-batch of transitions for further processing.
Utilizing the Bellman equation, the algorithm updates the $Q$-value,
performing a gradient descent step to minimize the loss in Eq. \eqref{eq:34-1}.
The neural network weights are then updated accordingly, and the final
learned weights that represent the policy learned are saved for future
use for Sim-to-Real implementation.

\subsubsection{Double DQN}

The previously mentioned $Q$-learning algorithm is known to overestimate
action values. In both $Q$-learning and DQN, the max operator in
\eqref{eq:303} utilizes identical values for both selecting and evaluating
an action. This raises the probability of selecting values that are
overestimated, resulting in overly optimistic estimations of value.
To address this issue, we can separate the selection process from
the evaluation. In the case of Double DQN \cite{van2016deep}, two
value functions are trained by randomly assigning experiences to update
either of the two value functions, yielding two sets of weights, $\theta$
and $\theta'$ . During each update, one set of weights is employed
to establish the greedy policy, while the other is utilized to assess
its value. This decoupling of selection and evaluation allows for
a clear comparison, aligning with the principles of $Q$-learning.

\subsubsection{Dueling DQN}

Starting with the standard DQN architecture, it is possible to divide
the network into two distinct streams: one dedicated to estimating
the state value, and the other focused on assessing state-dependent
action advantages. Following the two streams, the final module of
the network merges the outputs of state value and advantage, illustrated
in Figure \ref{fig:dueling}. The approach introduced in \cite{wang2016dueling}
suggests that the last module of the neural network executes the forward
mapping as follows:
\begin{alignat}{1}
	Q(s,a;\theta,\alpha,\beta)= & V(s;\theta,\beta)+(A(s,a;\theta,\alpha)\nonumber \\
	& \hspace{2.5em}-\frac{1}{|A|}\sum_{a'}A(s,a;\theta,\alpha))\label{eq:35}
\end{alignat}
The dueling architecture maintains an identical input-output interface
as the standard DQN architecture, ensuring an identical training process.
The Dueling DQN loss function can be expressed in terms of mean squared
error, such as:
\begin{equation}
	\mathcal{L}=\frac{1}{N}\sum_{i\in N}(Q_{j}-Q(s_{j},a_{j};\theta_{\text{prediction}}))^{2}\label{eq:36}
\end{equation}

\subsection{Policy based Agents}

\begin{algorithm}[h]
	\caption{PPO Docking learning Algorithm.}
	\label{algo:2}
	
	\textbf{Input:} $M$, $C$, $\gamma$, $\alpha_{\text{critic}}$,
	$\alpha_{\text{actor}}$, $\alpha_{w}$, $f_{p}$, and $\gamma_{w}$.
	
	\textbf{Output: $\theta$}
	
	\textbf{Initialization:}
	\begin{enumerate}
		\item[{\footnotesize{}1:}]  Initialize $\theta_{\text{actor}}$ and $\theta_{\text{critic}}$.
	\end{enumerate}
	\textbf{for} every episode\textbf{ $=1,\cdots,M$ do}
	\begin{enumerate}
		\item[{\footnotesize{}4:}]  Initialize $s_{1}.$
		\item[{\footnotesize{}5:}]  Compute $\vert X(\omega)\vert=\sqrt{Re(X(\omega))^{2}+X(f(\omega))}$
		from Eq. \eqref{eq:33-1} and \eqref{eq:34}.
		\item[{\footnotesize{}6:}]  Generate a random phase $\phi_{\omega}=[0,2\pi]$.
		\item[{\footnotesize{}7:}]  Generate the wave time signal from $z_{w}(t)=\mathcal{F}^{-1}[X(\omega)].$
	\end{enumerate}
	\textbf{\hspace{1cm}for }every time step \textbf{$k=1,\cdots,k_{f}$}
	\begin{enumerate}
		\item[{\footnotesize{}8:}]  Select $a_{k}$ according to old policy $\pi(s_{k}|\theta_{\text{old}})$.
		\item[{\footnotesize{}9:}]  Numerically integrate the dynamics in Eq. \eqref{eq:31} and \eqref{eq:32}
		using SciPy.
		\item[{\footnotesize{}10:}]  Store $(s_{k},a_{k},r_{k},s_{k+1})$ in $D$.
		\item[{\footnotesize{}11:}]  Calculate the estimates $A_{1},\cdots,A_{k_{f}}$ according to (put
		the equation here).
	\end{enumerate}
	\textbf{\hspace{1cm}end for}
	
	\textbf{\hspace{1cm}for }each epoch
	\begin{enumerate}
		\item[{\footnotesize{}12:}]  Sample a random mini batch of $L$ transitions from $D$.
		\item[{\footnotesize{}13:}]  execute a gradient descent step to minimize\\
		$\mathcal{L^{V}}=\hat{E}[(\hat{V}_{\theta_{\text{critic}}}(s_{k})-V_{\theta_{\text{critic}}}(s_{k}))^{2}]$.
		\item[{\footnotesize{}14:}]  execute a gradient descent step to minimize\\
		$\mathcal{L}^{\text{PPO}}=\hat{E}[\min(R_{k}(\theta_{\text{actor}}),clip(R_{k}(\theta_{\text{actor}}),1-\epsilon_{\text{PPO}},1+\epsilon_{\text{PPO}}))\hat{A_{k}}].$
		\item[{\footnotesize{}15:}]  update $\theta_{\text{critic}}\leftarrow\theta_{\text{critic}}+\alpha_{\text{critic}}\nabla_{\theta}J(\theta_{\text{critic}}).$
		\item[{\footnotesize{}16:}]  update $\theta_{\text{actor}}\leftarrow\theta_{\text{actor}}+\alpha_{\text{actor}}\nabla_{\theta}J(\theta_{\text{actor}}).$
	\end{enumerate}
	\textbf{\hspace{1cm}end for}
	
	\textbf{end for}
	\begin{enumerate}
		\item[{\footnotesize{}17:}] save $\theta_{\text{actor}}$
	\end{enumerate}
\end{algorithm}

\subsubsection{Proximal Policy Optimization (PPO)}

The PPO algorithm, introduced in \cite{schulman2017proximal}, is
a policy gradient method applicable to environments featuring both
discrete and continuous action spaces. It trains a stochastic policy
using the on-policy actor-critic method. The actor maps observations
into actions, while the critic provides an expected reward for the
given observation. Initially, it gathers a set of trajectories per
epoch by sampling from the most recent version of the stochastic policy.
The policy undergoes updates through SGD optimization, whereas the
value function is adjusted using a gradient descent algorithm.

\begin{figure*}[h]
	\begin{centering}
		\includegraphics[scale=0.55]{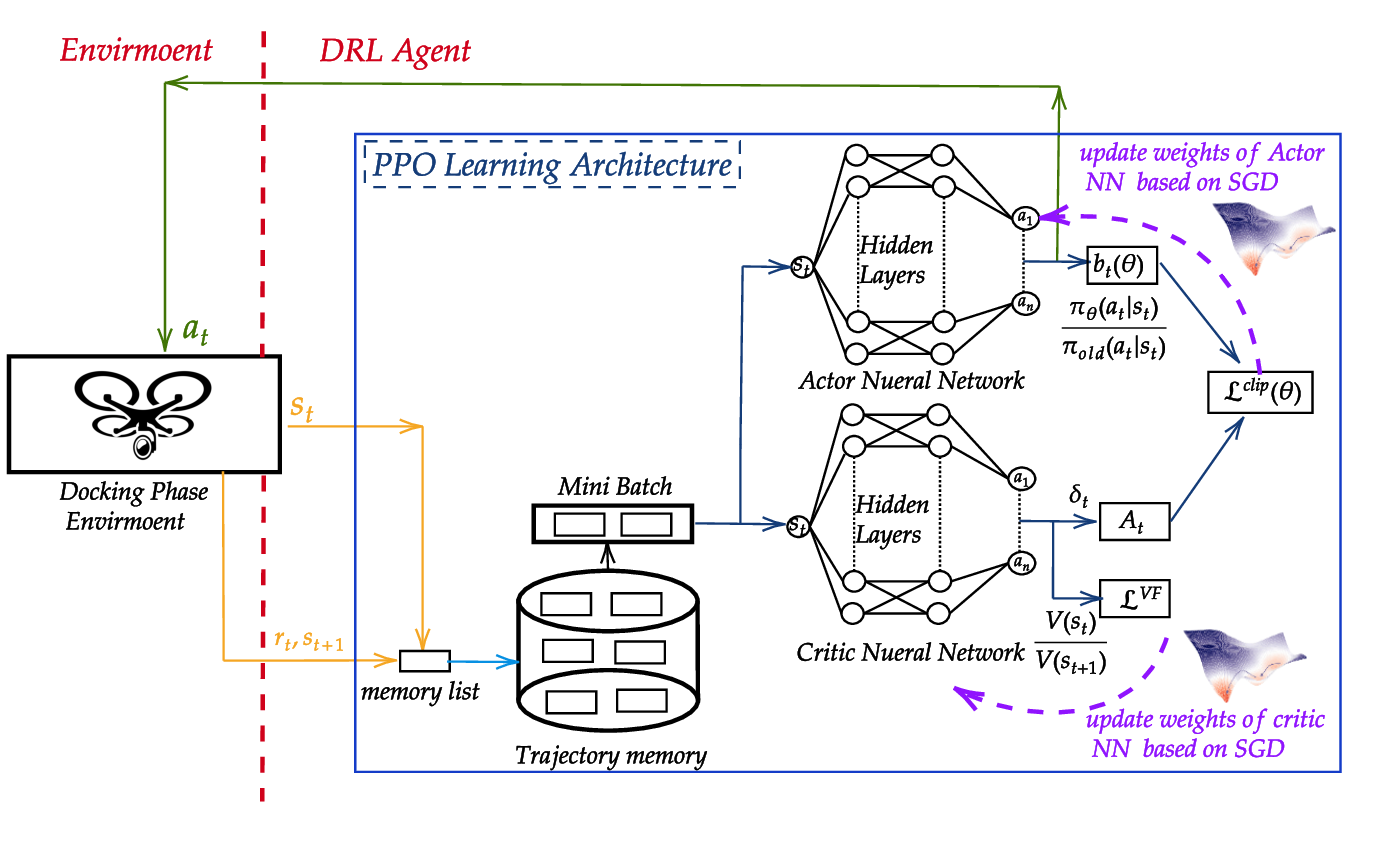}
		\par\end{centering}
	\caption{PPO Agent architecture interacting with the docking phase environment.}
\end{figure*}
Define $\delta_{k}=r_{k}+\gamma V(s_{k+1})-V(s_{k}).$ The Generalized
Advantage Estimator (GAE) can be computed as follows \cite{schulman2015high}:
\begin{equation}
	\hat{A_{k}}=\delta_{k}+(\gamma\lambda)\text{\ensuremath{\delta_{k+1}+(\gamma\lambda)^{2}\delta_{k+2}+\ldots+(\gamma\lambda)^{\mu-k+1}\delta_{\mu-1}.}}\label{eq:38}
\end{equation}
where $\mu$ is the sampled mini-batch size and the loss function
of the policy gradient can be expressed as
\begin{equation}
	\mathcal{L}=\hat{E}[|\hat{V}_{k}^{\text{target}}(s_{k})-V_{k}(s_{k})|]\label{eq:39}
\end{equation}
where $\hat{V}_{k}^{\text{target}}(s_{k})=r_{k+1}+\gamma V_{k}(s_{k+1})$
and the weights of the policy neural network will be updated by an
SGD algorithm. In \cite{pmlr-v37-schulman15} Trust Region Policy
(TRP) utilized the loss function with respect to the following constraint:
\begin{equation}
	\hat{E}[KL[\pi_{\text{old}}(\cdotp|s_{k}),\pi_{\theta}(\cdotp|s_{k})]]\leq\delta_{0}\label{eq:40}
\end{equation}
where $\delta_{0}$ is a small number and $KL$ is the kullback-leibler
divergence defined in \cite{pmlr-v37-schulman15}. The PPO algorithm
in \cite{schulman2017proximal} has the same idea as the TRP but it
is much simpler to be implemented with better sample efficiency. Define
the probability ratio $R_{k}(\theta)=\frac{\pi_{\theta}(a_{k}|s_{k})}{\pi_{\text{old}}(a_{k}|s_{k})}$.
The PPO loss function is given by 
\begin{multline}
	\mathcal{L}^{\text{PPO}}(\theta_{\text{actor}})=\hat{E}[\min(R_{k}(\theta_{\text{actor}}),\\
	\text{clip}(R_{k}(\theta_{\text{actor}}),1-\epsilon_{\text{PPO}},1+\epsilon_{\text{PPO}}))\hat{A_{k}}]\label{eq:41}
\end{multline}
where $\epsilon$ is the clipping parameter. The clipped objective
function $\mathcal{L}^{\text{PPO}}$ prevents PPO from exhibiting
a bias towards actions with positive advantages and enables it to
promptly steer clear of actions associated with a negative advantage
function within a mini-batch of samples. The parameters of $\pi_{\theta}$
are updated by an SGD algorithm with the gradient $\triangle\mathcal{L}^{\text{PPO}}$.
The update of the critic and actor weights is done using SGD as follows:

\begin{equation}
	\theta_{\text{actor}}=\theta_{\text{actor}}-\alpha_{\text{actor}}\nabla\mathcal{L}^{\text{PPO}}(\theta_{\text{actor}})\label{eq:42}
\end{equation}

\begin{equation}
	\theta_{\text{critic}}=\theta_{\text{critic}}-\alpha_{\text{critic}}\nabla\mathcal{L}^{\text{PPO}}(\theta_{\text{critic}})\label{eq:43}
\end{equation}
where $\alpha_{\text{actor}}$ and $\alpha_{\text{critic}}$ are the
learning rate of the actor and critic respectively. 

\paragraph*{Algorithm \ref{algo:2} Discussion}

The algorithm begins by initializing the replay memory with capacity
$C$, number of episodes $M$, discount factor $\gamma$, learning
rate of the actor and critic neural networks $\alpha_{\text{critic}}$,
$\alpha_{\text{actor}}$, and other parameters identifying the wave
spectral energy in Eq. \eqref{eq:33-1} as $\alpha_{w}$, $f_{p}$,
and $\gamma_{w}$. In steps 4-7, the random wave in the time domain
is generated within each epoch as shown in Eq. \eqref{eq:33-1} and
\eqref{eq:34}. For each time step, it samples a random mini-batch
of transitions from the memory and updates the Critic network by minimizing
the value function loss and the actor-network using the PPO loss described
in Eq. \eqref{eq:41}. Both networks (actor and critic model) are
updated using SGD with the learning rates $\alpha_{\text{critic}}$
and $\alpha_{\text{actor}}$. The final weights of the actor-network
are saved as $\theta_{\text{actor}}$ representing the policy learned. 

\section{Numerical Results \label{sec:Numerical-Experiments}}
\begin{table*}[h]
	\caption{\label{Table1}Agent Parameters.}
	
	\centering{}%
	\begin{tabular}{ccccc}
		\toprule 
		Hyper parameters & DQN & Double DQN & Dueling DQN & PPO\tabularnewline
		\midrule
		\midrule 
		Learning Rate (Both) & 0.00001 & 0.00001 & 0.00001 & 0.0003\tabularnewline
		\midrule 
		Batch size (Both) & 64 & 64 & 64 & 5\tabularnewline
		\midrule 
		Discount (Both) & 0.995 & 0.995 & 0.995 & 0.99\tabularnewline
		\midrule 
		Memory (Both) & 10000 & 10000 & 10000 & 10000\tabularnewline
		\midrule 
		Target Syn Frequency (DQNs) & 10 & 10 & 10 & -\tabularnewline
		\midrule 
		Soft Update Weight (DQNs) & 0.8 & 0.8 & 0.8 & -\tabularnewline
		\midrule 
		Gradient Clipping Value (DQNs) & 1 & 1 & 1 & -\tabularnewline
		\midrule 
		Learning warm start (DQNs) & 200 & 200 & 200 & -\tabularnewline
		\midrule 
		Likelihood Ratio Clipping (PPO) & - & - & - & 0.2\tabularnewline
		\midrule 
		Learning Frequency (PPO) & - & - & - & 20\tabularnewline
		\midrule 
		Learning Epochs (PPO) & - & - & - & 4\tabularnewline
		\midrule 
		GAE Lambda value (PPO) & - & - & - & 0.95\tabularnewline
		\bottomrule
	\end{tabular}
\end{table*}
\begin{table*}[h]
	\caption{\label{table2}Neural Network Parameters.}
	
	\centering{}%
	\begin{tabular}{ccccc}
		\toprule 
		Network shape & DQN & Double DQN & Dueling DQN & PPO\tabularnewline
		\midrule
		\midrule 
		Input Layer Neurons & 1 & 1 & 1 & 1\tabularnewline
		\midrule 
		Input Layer Activation Function & Tanh & Tanh & Tanh & Tanh\tabularnewline
		\midrule 
		First Hidden Layer Activation Neurons & 32 & 32 & 32 & 32\tabularnewline
		\midrule 
		Second Hidden Layer Activation Function & Tanh & Tanh & Tanh & Tanh\tabularnewline
		\midrule 
		Second Hidden Layer Neurons & 32 & 32 & 32 & 32\tabularnewline
		\midrule 
		Third Hidden Layer Activation Function & Tanh & Tanh & Tanh & Tanh\tabularnewline
		\midrule 
		Third Hidden Layer Neurons & 16 & 16 & 16 & 16\tabularnewline
		\midrule 
		Output Layer Neurons & 3 & 3 & 3 & 1\tabularnewline
		\midrule 
		Output Layer Activation Function & - & - & - & Softmax\tabularnewline
		\bottomrule
	\end{tabular}
\end{table*}

In this section, the main results will be discussed. The computing
source used to carry out simulations in this section has the following
specifications: Windows 11 (64-bit processor), CPU (Intel(R) i7-10760H,
2.60GHz, 2592 MHz, 6 core, and 12 logic processors), and 16 GB RAM.
\begin{figure}[h!]
	\begin{centering}
		\includegraphics[scale=0.32]{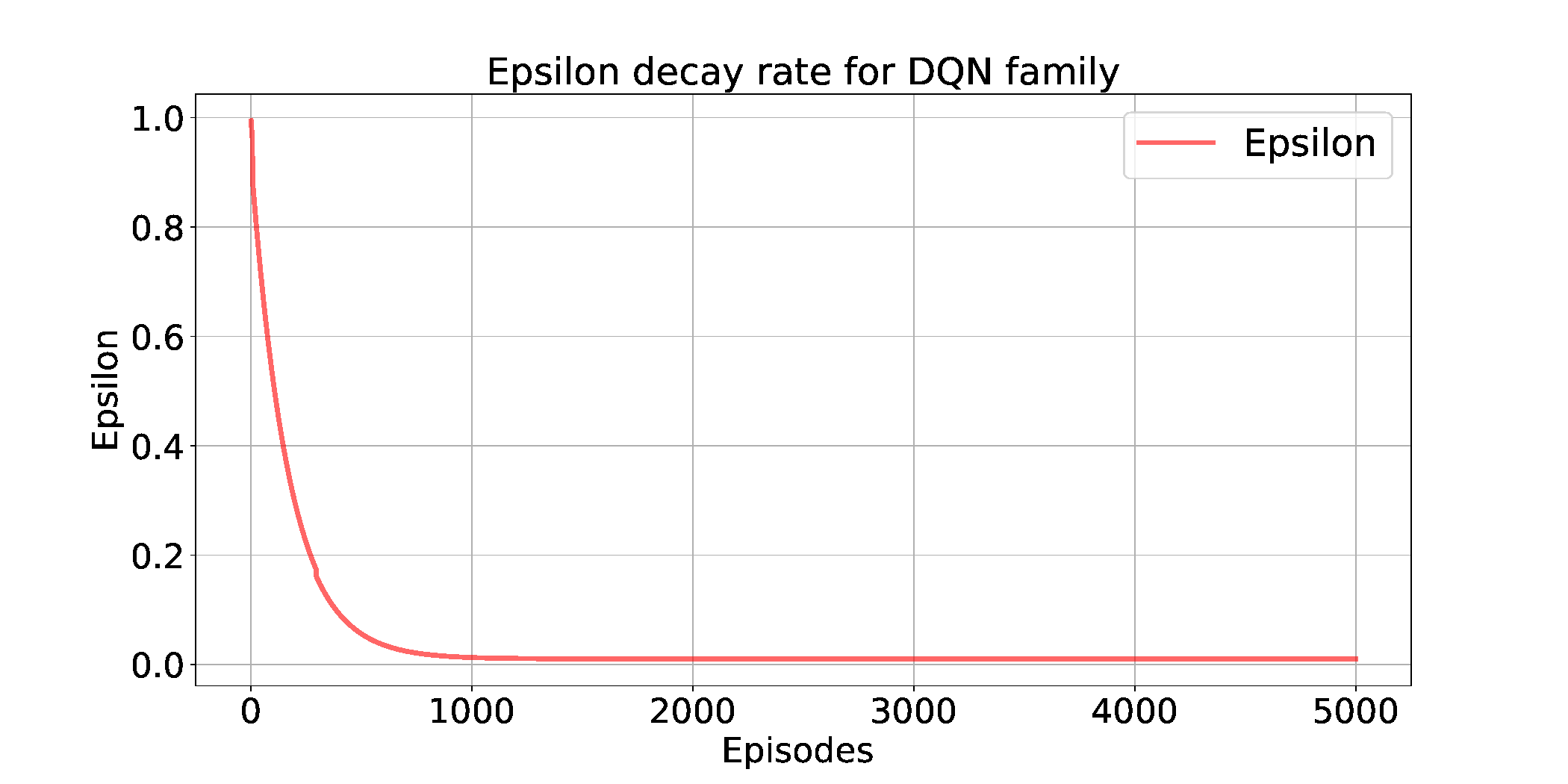}
		\par\end{centering}
	\caption{The $\epsilon$ greedy value used in DQN agents. The initial and final
		greedy values were $\epsilon_{0}=1$ $\epsilon_{f}=0.05$ respectively. }
	
	\label{fig:eplsion}
\end{figure}
\begin{figure}[h!]
	\begin{centering}
		\includegraphics[scale=0.27]{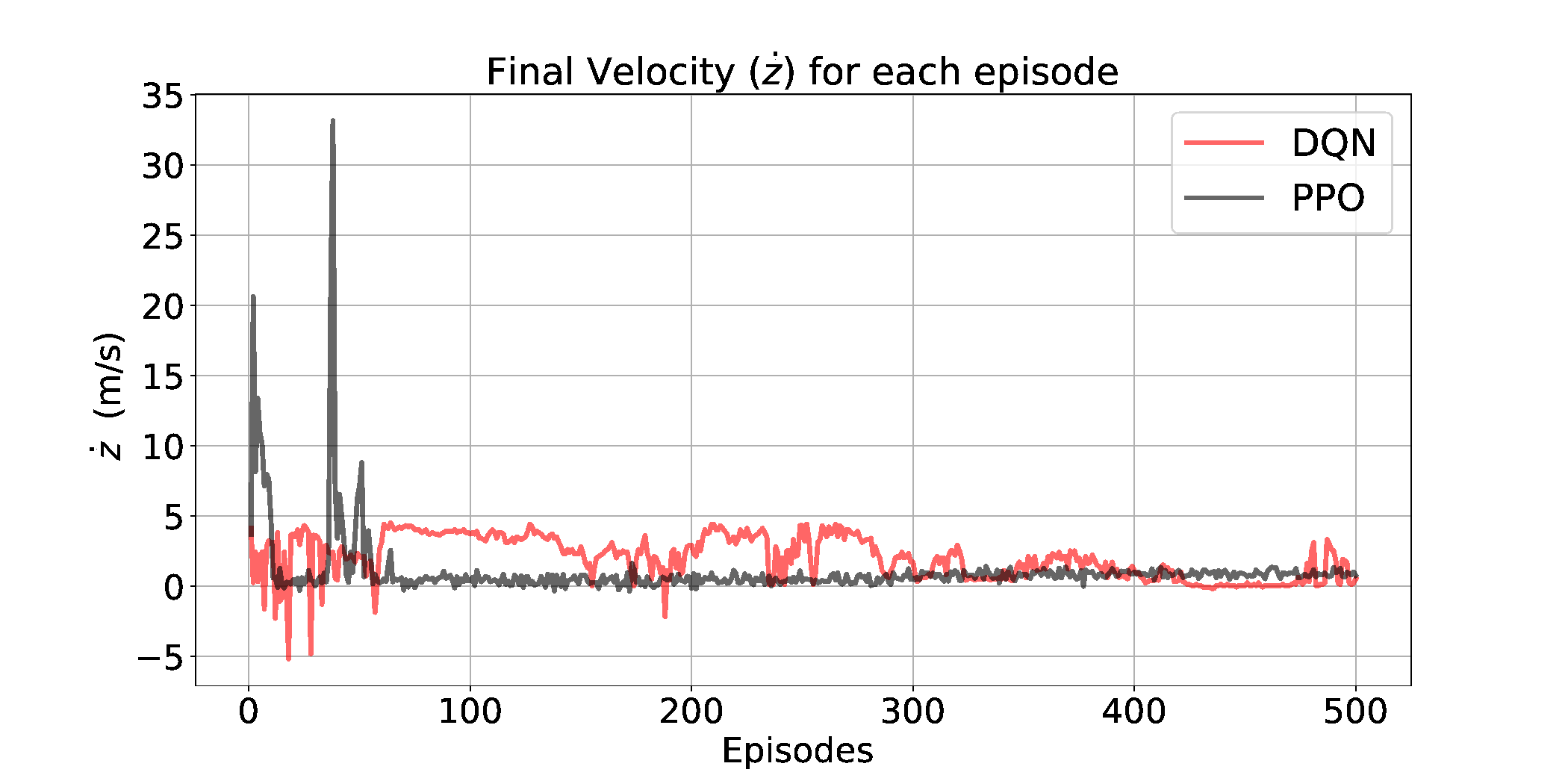}
		\par\end{centering}
	\caption{A comparison between the final Impact Velocity $\dot{z}$ of the VTOL-UAV
		in the case of PPO and DQN agents.}
	
	\label{fig:final-velcoity}
\end{figure}
\begin{figure*}[h!]
	\begin{centering}
		\includegraphics[scale=0.41]{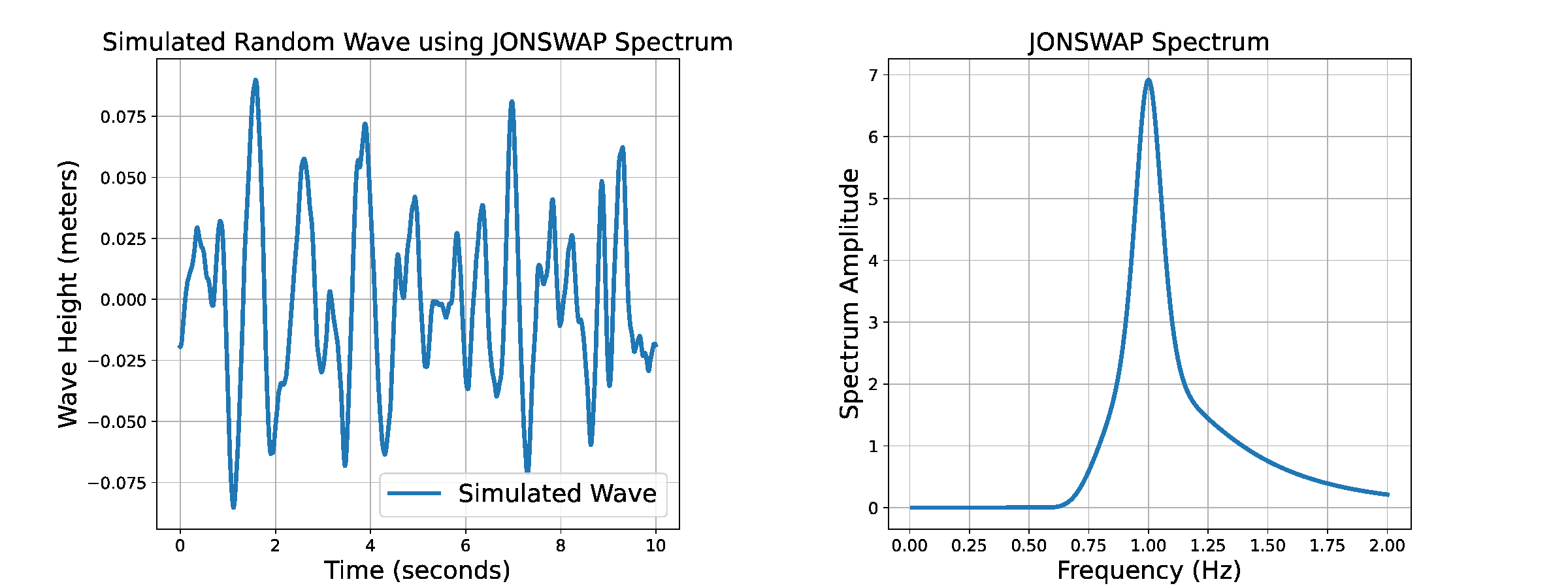}
		\par\end{centering}
	\caption{The left portion of the figure depicts one instance of the simulated
		random wave using the JONSAWP model. At each episode, a new randomized
		simulated wave is generated to enhance the generalization of the policy
		for sim-to-real transfer. The right portion shows the energy spectrum
		of the same simulated wave.}
	
	\label{fig:JONSWAP}
\end{figure*}
\begin{figure*}[h!]
	\begin{centering}
		\includegraphics[scale=0.38]{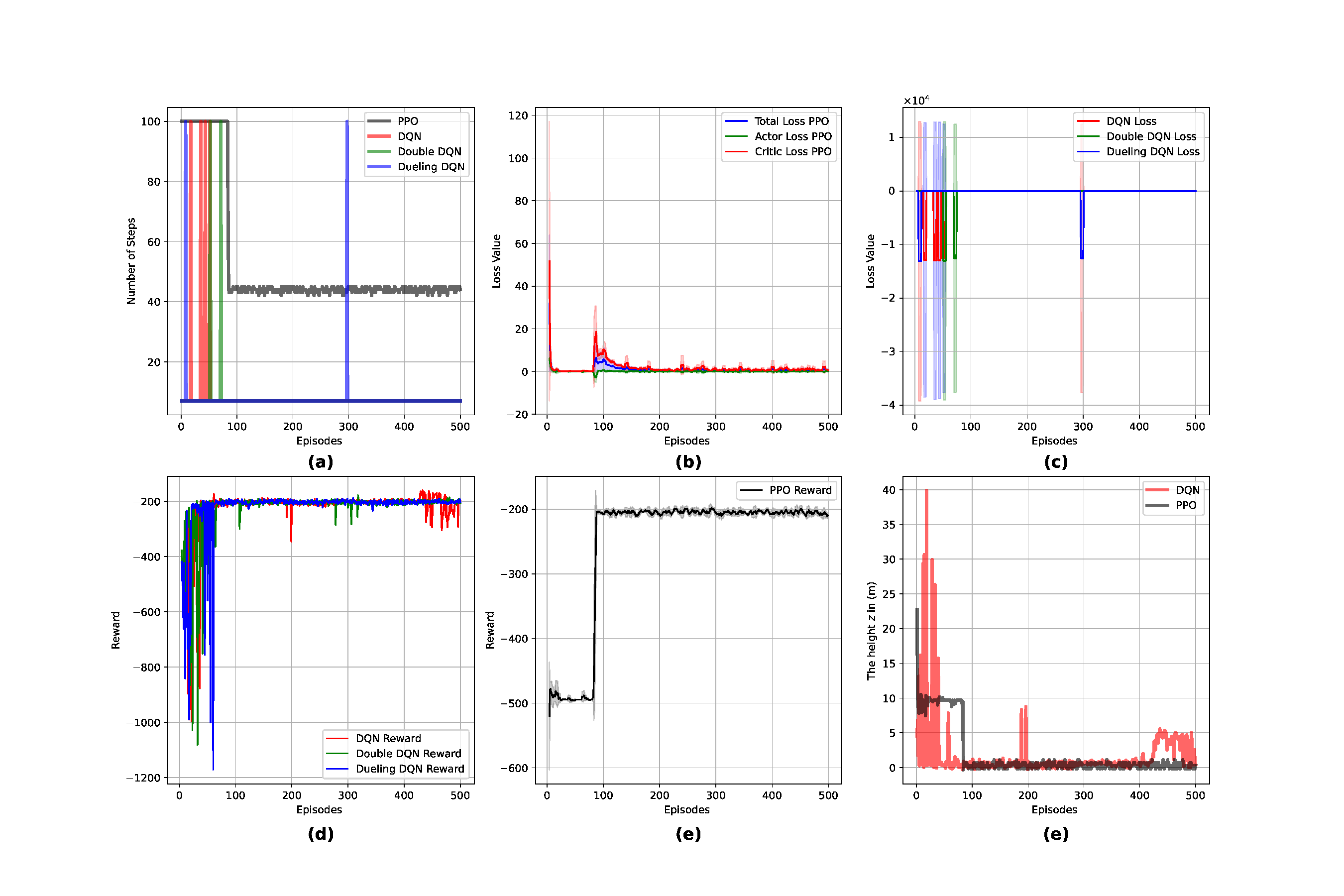}
		\par\end{centering}
	\caption{\textcolor{black}{Part (a) compares the number of time steps needed
			to land for all the agents. Part (b) illustrates the actor and critic
			loss moving average value of the PPO agent, while part (c) shows the
			moving average loss value of the DQNs agents. Part (d) and (e) depict
			the moving average of the reward of DQNs and PPO agents respectively.
			Finally, part (f) compares the final height achieved by DQN and PPO
			agents. The shaded parts in all figures represent the standard deviation
			of the moving average.}}
	
	\label{fig:Results}
\end{figure*}
The $\text{\ensuremath{\epsilon}}$ greedy value used in all DQNs
agents started from $\epsilon_{0}=1$ and decaying exponentially until
$\epsilon=0.01$. Figure \ref{fig:eplsion} illustrates the epsilon
$\epsilon$ greedy values utilized in the DQN agents. The initial
and final greedy values were $\epsilon_{0}=1$ and $\epsilon_{f}=0.05$,
respectively, allowing 80 episodes of exploration and 420 episodes
for exploitation. Figure \ref{fig:JONSWAP} depicts one instance of
the simulated random wave and the correspondent energy spectrum using
the JONSAWP model. At each episode, a new randomized simulated wave
is generated to enhance the generalization of the policy for sim-to-real
transfer. The learning rates $\alpha$ used to update the policy of
the DQN agents was $\alpha=0.0001$ based on the gradient of the loss
function in Eq. \eqref{eq:34-1} and \eqref{eq:36}. The learning
rates $\alpha_{\text{actor}}$ and $\alpha_{\text{critic}}$ used
to update the policy based on the loss function in Eq. \eqref{eq:41}
was $\alpha_{\text{actor}}=\alpha_{\text{critic}}=0.0003$. A gradient
clipping value of 1 is used in the case of the DQN, to guarantee a
stable performance of the DQN agents. A likelihood ratio clipping
of 0.2 was used for the PPO agent with a learning Frequency of 20.
The shape of the neural networks in all the agents is kept the same
for the sake of fair comparison between the agents. The neural networks
contained three layers with 32, 32, and 16 neurons. The hyperbolic
tan function was utilized as the activation layer for all the layers.
The remaining hyperparameters and neural network structure are shown
in Table \ref{Table1} and Table \ref{table2}.

Figure \ref{fig:final-velcoity} illustrates the difference between
the final impact velocity in the case of PPO (shown in black) and
the DQN agent (shown in red). The PPO agent was able to learn better
policy to track the required velocity $\dot{z}_{d}$. An exponential
decay function was utilized to represent the required velocity $\dot{z}_{2d}$.
Figure \ref{fig:Results}(a) compares the number of time steps needed
to land between DQN and PPO agents. The number of steps needed by
the PPO agent to land is lower than the DQN agents. The PPO agent
was able to learn more efficient policies that require less steps
to land compared to the DQN agents. Figure \ref{fig:Results}(b) illustrates
the actor and critic loss moving average value of the PPO agent. The
PPO agent showed smooth learning capabilities, where both actor and
critic values converge to zero in less than 200 episodes. Figure \ref{fig:Results}(c)
depicts the moving average value of the DQN agents for 500 episodes.
Unlike the PPO, the moving average loss value of the DQN agents suffered
from strong fluctuations, especially during the exploration phase.
Figure \ref{fig:Results}(d) illustrates the moving average reward
of the DQN agents. The DQN agent suffered from the forgetting phenomenon,
where the agent forgot the optimal policy at the end of the training.
This phenomenon is usually related to overfitting. The Double DQN
showed better performance compared to the normal and Dueling DQN.
Figure \ref{fig:Results}(d) depicts the moving average reward value
of the PPO agent. The PPO agent average reward converged to a value
around $-200$ similar to the DQN agents. We experienced less sensitivity
of the hyperparameters in the case of the PPO agent compared to the
DQN. The performances of the agents have been evaluated in Table \ref{table3}
in terms of final impact velocity, the time needed to land, and the
inference time (computational time to recall the solution). The PPO
agent was able to achieve the least value of final impact velocity
due to its complicated and efficient policy compared to the DQNs.
Due to its complicated policy, the PPO agent has the longest time
to land compared to the DQN agents with 4.9 seconds. 

\begin{table}[h]
	\caption{\label{table3}Evaluation of the trained agents.}
	
	\centering{}%
	\begin{tabular}{>{\centering}p{1.5cm}>{\centering}p{1.5cm}>{\centering}p{1.5cm}>{\centering}p{1.5cm}}
		\toprule 
		& Impact Velocity (m\textbackslash s) & Time to Land (s) & Inference Time (ms)\tabularnewline
		\midrule
		\midrule 
		PPO & 0.327 & \textbf{4.9} & \textbf{6.788}\tabularnewline
		\midrule 
		DQN & 0.820 & 5.4 & 9.854\tabularnewline
		\midrule 
		Double DQN & \textbf{0.223} & 7.7 & 7.220\tabularnewline
		\midrule 
		Dueling DQN & 2.419 & 6.3 & 11.296\tabularnewline
		\bottomrule
	\end{tabular}
\end{table}

\section{Conclusions \& Future Work \label{sec:Conclusions-=000026-Future}}

A novel reinforcement learning approaches for sim-to-real policy transfer
for Vertical Take-Off and Landing Unmanned Aerial Vehicle (VTOL-UAV)
designed for landing on offshore docking stations in maritime operations
has been proposed. The Joint North Sea Wave Project (JONSWAP) spectrum
model has been employed to create a wave model for each episode, enhancing
policy generalization for sim-to-real transfer. A set of DRL algorithms
has been tested through numerical simulations including value-based
agents such as Deep \textit{Q} Networks (DQN) and policy-based agents
such as Proximal Policy Optimization (PPO). The PPO
agent achieved more stable and efficient learning to land in uncertain
environments, which in turn enhances the likelihood of successful
Sim-to-Real transfer. The simulation showed the capability of PPO
to learn more complicated policies which turned out to be more efficient
than the DQN policies. The learned policy can be utilized in real-time
for control in the landing phase, where the policy will map the current
state to optimal control action which has been learned through offline
experience. For future work, exploring how agents perform with visual
feedback from the on-board camera could be an intriguing area of investigation.

\section*{Acknowledgments}

This work was supported in part by the National Sciences and Engineering Research Council of Canada (NSERC), under the grants RGPIN-2022-04937.

\balance
\bibliographystyle{IEEEtran}
\bibliography{dockingref}
% name your BibTeX data base
		
	\end{document}